\documentclass[letterpaper, 10 pt, conference]{ieeeconf}  

\IEEEoverridecommandlockouts                              

\usepackage[pdftex]{graphicx} 
\usepackage{times} 
\usepackage{amsmath} 
\usepackage{amssymb}  
\usepackage{bm}
\usepackage[nodisplayskipstretch]{setspace}
\usepackage[caption=false,font=footnotesize]{subfig}
\usepackage{multirow}
\usepackage[T1]{fontenc}
\title{\LARGE \bf
Learning of Multi-Context Models for Autonomous Underwater Vehicles
}

\author{Bilal Wehbe$^{1, 2}$,  Octavio Arriaga$^{1}$, Mario
Michael Krell$^{2}$, and Frank Kirchner$^{1, 2}$
\thanks{$^{1}$ DFKI - Robotic Innovation Center, Bremen, Germany. \newline
{\tt\small \{first name.last name\}@dfki.de}}%
\thanks{$^{2}$ Robotics Research Group, University of Bremen, Germany.}}

\begin{document}

\maketitle
\thispagestyle{empty}
\pagestyle{empty}

\begin{abstract}
Multi-context model learning is crucial for marine robotics where several factors
can cause disturbances to the system's dynamics. This work addresses the
problem of identifying multiple contexts of an AUV model. We build a simulation model
of the robot from experimental data, and use it to fill in the missing data and generate different
model contexts. We implement an architecture based on long-short-term-memory (LSTM) networks
to learn the different contexts directly from the data. We show that the LSTM network can achieve
high classification accuracy compared to baseline methods, showing robustness against noise and
scaling efficiently on large datasets.
\end{abstract}

\section{INTRODUCTION}
A robotic model is an essential tool for control,
action and path planning, and several other  applications. Classically,
models were manually engineered by humans for specific robotic designs and
applications, which restrict their usability when the environmental conditions
or the robot mechanics are non-stationary. This issue as well presents itself
critically in marine robotics, where robots have to operate persistently in harsh
and unpredictable environments for weeks or even months.
Machine learning methods can avoid the manual hand
crafting of robotic models, and instead learn these models directly from the
data streams acquired by the robot during operation.
Furthermore, machine learning can generalize better on larger state space of the model and
take into account nonlinearities that are most of the times neglected by
classical physics-based approaches \cite{nguyen2011model}. 

Model learning has been proven to be an efficient methodology in various robotic
domains such as inverse kinematics and dynamics control, robot manipulation,
locomotion or navigation. However learning these models may not always be
straightforward and is still being faced with several challenges \cite{nguyen2011model}.
For most applications, model learning is regarded as a regression problem mapping
the robot's states and actions.  However, in cases when the robot's dynamics or
operating environment are  non-stationary, standard regression techniques cannot
be used since they are not able to represent the full state space of the model. This
problem is commonly known as multi-context learning
\cite{jamone2014incremental}. The common challenges that face
multi-context learning is discovering the correct number of contexts present in
the data space, and identifying the current context the robot is in at a certain
time. Furthermore, the incomplete state space of the sampled data, poses one of the 
major problems for learning algorithms. Generally any learning method
requires a large and rich enough dataset to be able to generalize properly. In such
cases, learning from simulation can help improve the model by providing an
alternative for missing data.

\begin{figure}[t!]
\centering
\includegraphics[trim=0 0 0 0,clip,width=0.43\textwidth]{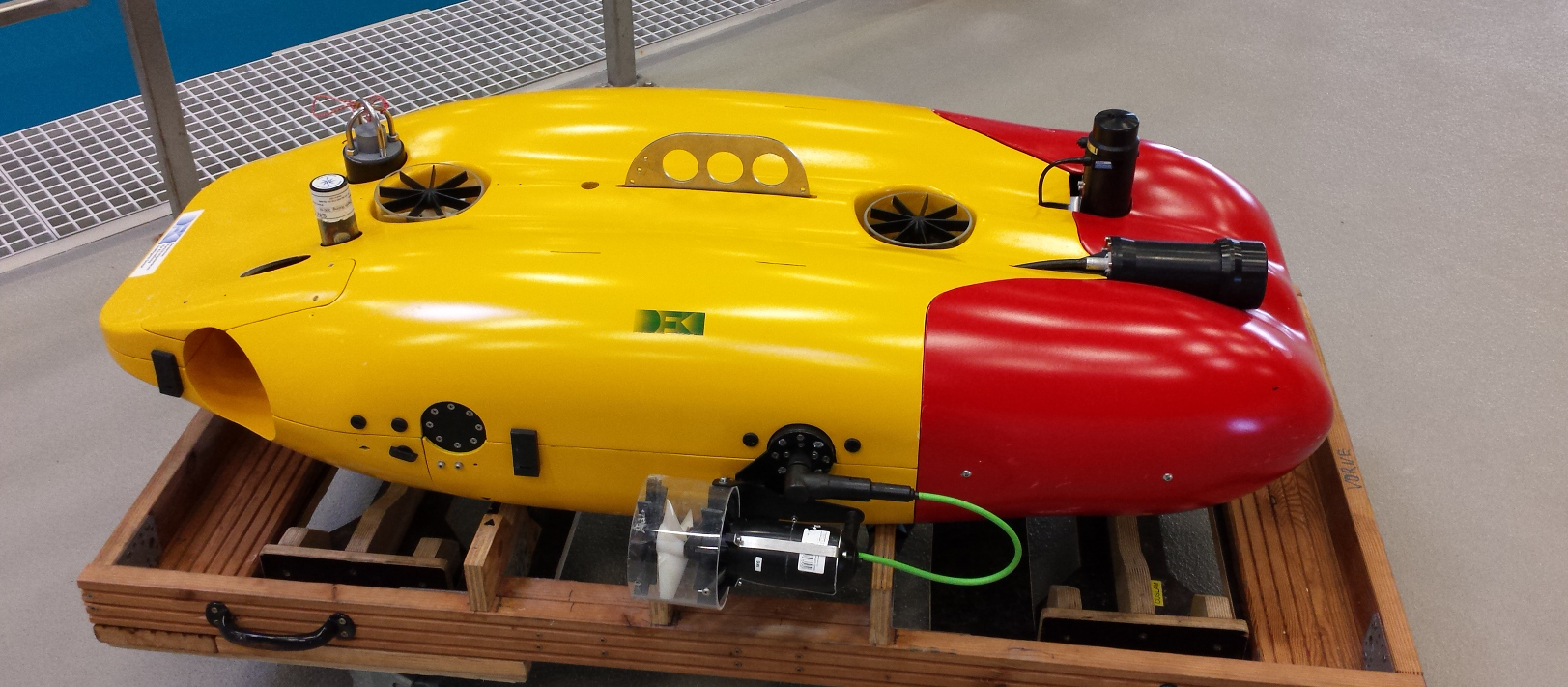}
\caption{The AUV Dagon used in our experiment.}
\label{fig:dagon}
\vspace{-0.3cm}
\end{figure}

In this work, we study the case of an autonomous underwater vehicle (AUV) where
many factors such as salinity or density fluctuations, biofouling, or body damage
can cause a change of the robot's model context. We aim to classify the
different contexts of an AUV model resulting from disturbances in its dynamics. First,
we generate a simulation model of the AUV Dagon (Fig.~\ref{fig:dagon}) that is
learned from real data collected from experiments. We then induce several
faults into the simulated model and generate a sufficiently rich dataset which
contains different model contexts and covers a large area of the model's
state space. We regard the multi-context learning as a multi-class classification
problem, where each context of the robot model is assigned to a unique label.
Using the generated dataset, we build a gating network that classifies the correct
model context seen by the robot at a certain time. We build the gating network using 
a long-short-term-memory (LSTM); therefore, modeling the data as a time series.
We show that the LSTM network can achieve a better performance when compared to standard
classification methods such as support vector machines (SVM), random forests (RF), and
multi-layer perceptrons (MLP).

\subsection{Related Work}
Multi-context learning has been recently a quite active topic for robotic model
learning, most frequently in the area of inverse dynamics modeling for 
manipulator arms \cite{jamone2014incremental,
calandra2015learning}. The mixture of experts (ME) approach
\cite{yuksel2012twenty} is a frequently used method  for
learning multi-context models, where the data is clustered into smaller groups
and subsequently a local model (or an expert) is built for each cluster.  An infinite
mixture of linear experts was used in \cite{jamone2014incremental} to capture
the different contexts of an inverse dynamic model for a humanoid arm manipulating
objects with different weights. However, since the linear experts only model the
system locally, this causes the number of experts to increase quickly as the
system perceives new contexts. This method results in the number of experts not
representing the correct number of contexts, for example around 60 experts 
were needed to represent only two contexts in \cite{jamone2014incremental}.
A mixture of Gaussian process (GP) experts was used in \cite{calandra2015learning}
to model different contact models for a humanoid manipulator arm. An SVM
was used as a gating network to select between the 
different contexts, which achieved an accuracy comparable to a manually tuned
heuristic method described in the same paper. A multi-context model of a wheeled
mobile robot was learned using an infinite mixture of Gaussian process experts
in \cite{mckinnon2017learning}. Here, a Dirichlet process (DP) gating network was
used instead, which allows the classification of different contexts in an
unsupervised manner. One disadvantage of this method is that clustering via the
DP may not always predict the true context, but can easily get confused depending
on the density distribution of the training data. For example, two batches from
the same context  would be classified as two different contexts if they have different
densities. Another drawback is that all training samples have to be
used for a prediction.

While GPs are the state of the art methods for robotic model learning, it is a
well known fact that their computational complexity scales cubically with
the number of training samples $O(n^3)$. This fact makes GPs unlikely to benefit
from having big datasets that cover larger regions of the model's state space.
A recent study \cite{Humanoids2017Rueckert} shows that LSTM
networks with a training time complexity of only $O(n)$, can outperform GPs for learning
the inverse dynamics of a manipulator arm. The fact that LSTMs exploit the
temporal correlations in the data makes this method suitable for learning dynamic
models. Furthermore, their good computational efficiency and their scalability
on big datasets is a major advantage for long-term learning settings.

In the context of marine robotics, the application of machine learning is still
relatively scarce. For instance, convolutional neural networks were used for sonar image
recognition in \cite{valdenegro2017best}, moreover, the classification of autonomous
underwater vehicle (AUVs) trajectories was studied in \cite{alvarez2017}. Yet
model learning for underwater vehicles is still understudied in this field. Locally
weighted projection regression was used to identify the mismatch between
the physics based model and the data output from the vehicle's navigation
sensors in \cite{fagogenis2014improving}. In \cite{shafiei2015application},
a nonlinear auto-regressive network with a gating network based on a genetic
algorithm was used to identify a simulated model of an AUV with variable mass.
In previous work \cite{wehbe2017learning,wehbe2017online} we showed that
support vector regression can be a good candidate for learning AUV dynamic
models.

In this work, we build upon our previous findings, focusing our efforts on finding
a good candidate for a gating network which can provide an accurate classification
of different model contexts while at the same time being able to scale on big
datasets.

\section{Problem Statement}
We start with stating the formulation of the dynamic model of AUVs first. Next, we
describe the functionality of the gating network as a classifier for different model
contexts, followed by a description of the methods being evaluated.
\subsection{The Dynamic Model}
Following the notation of \cite{fossen2002marine}, the nonlinear 6-degrees-of-freedom
(DOF) equations of motion of an underwater vehicle are generally described by 
\begin{equation}
 \bm{\dot{\eta}} = J(\bm{\eta})\bm{\nu}\, ,\\
  \label{kinematic}
\end{equation}
\vspace{-0.5cm}
\begin{equation}
 M\bm{\dot{\nu}}+C(\bm{\nu})\bm{\nu}+d(\bm{\nu})+g(\bm{\eta})=
 \tau + \zeta(\bm{\eta},\bm{\nu},\bm{\tau})\, .
 \label{dynamic}
\end{equation}
Eq.~(\ref{kinematic}) represents the kinematics equation, where
$\bm{\eta} = [x \ y \ z \ \phi\ \theta \ \psi]^T$ is the pose of the vehicle
in a fixed coordinate frame,
and $\bm{\nu} = [u \ v \ w \ p \ q \ r]^T$ is the velocity of the vehicle expressed
in a body-attached frame. The term $J(\bm{\eta}) \in \mathbb{R}^{6\times 6}$ 
represents the nonlinear Euler angles transformation matrix. The dynamic equations
of motion are represented in Eq.~(\ref{dynamic}), where $M$ is the total mass
(dry + added mass) of the submerged vehicle, and $C(\bm{\nu})$ is the 
Coriolis and centripetal forces and moments. $d(\nu)$ represents the hydrodynamic
damping, and $g(\bm{\eta})$ accounts for the buoyancy and gravitational efforts. $\tau$
is a vector denoting the actuators forces and moments, and $\zeta$ is a term
representing all unmodeled dynamics and sensor noise. We follow the definition of
\cite{fossen2002marine} for all terms except for the damping term
$d(\bm{\nu})$, where we use the formulation of McFarland and Whitcomb
\cite{mcfarland2013comparative},
\begin{equation}
d(\bm{\nu}) = \left(\sum_{i=1}^{6}|\bm{\nu_i}| D_i \right)\bm{\nu} \, ,
\label{drag}
\end{equation}
where $D_i \in \mathbb{R}^{6\times 6},\. i=1,...,6$ are six matrices representing
the fully coupled quadratic damping. The choice of this damping model is based
upon our findings in \cite{wehbe2017learning}, where we showed that the 
McFarland-Whitcomb model achieves a decent performance amongst physics-based
models.
We rewrite Eq.~(\ref{dynamic}) as a forward model denoted by
\begin{equation}\label{model}
 \bm{\dot{\nu}} = \mathcal{F}(\bm{\eta},\bm{\nu},\bm{\tau})\,,\\
\end{equation}
where $ \mathcal{F}$ represents the dynamics function resulting from rearranging 
Eq.~(\ref{dynamic}). 

We define a context $(\bm{c})$ of the model as a unique set of the hydrodynamic
parameters underlying the function $\mathcal{F}^{\bm{c}}$. In other words, any
changes in the set of the model parameters (mass, coriolis, damping, buoyancy, ...)
will result in a different model context. For practicality reasons, we will constrain
the set of possible contexts to a finite set $\{\mathcal{F}^{\bm{c}}, \, \bm{c}=1,...,n\}$.
Accordingly, we will denote a dataset sampled from a model context $\mathcal{F}^{\bm{c}}$
as the vector formed by the model's states and control inputs,
$\mathfrak{D}^{\bm{c}} = \left\langle \bm{\dot{\nu}},\bm{\nu}, \bm{\eta},
\bm{\tau} \right\rangle ^{\bm{c}}$.
\subsection{The Gating Network}
\begin{figure}[t!]
\centering
\includegraphics[trim=0 0 0 0,clip,width=0.41\textwidth]{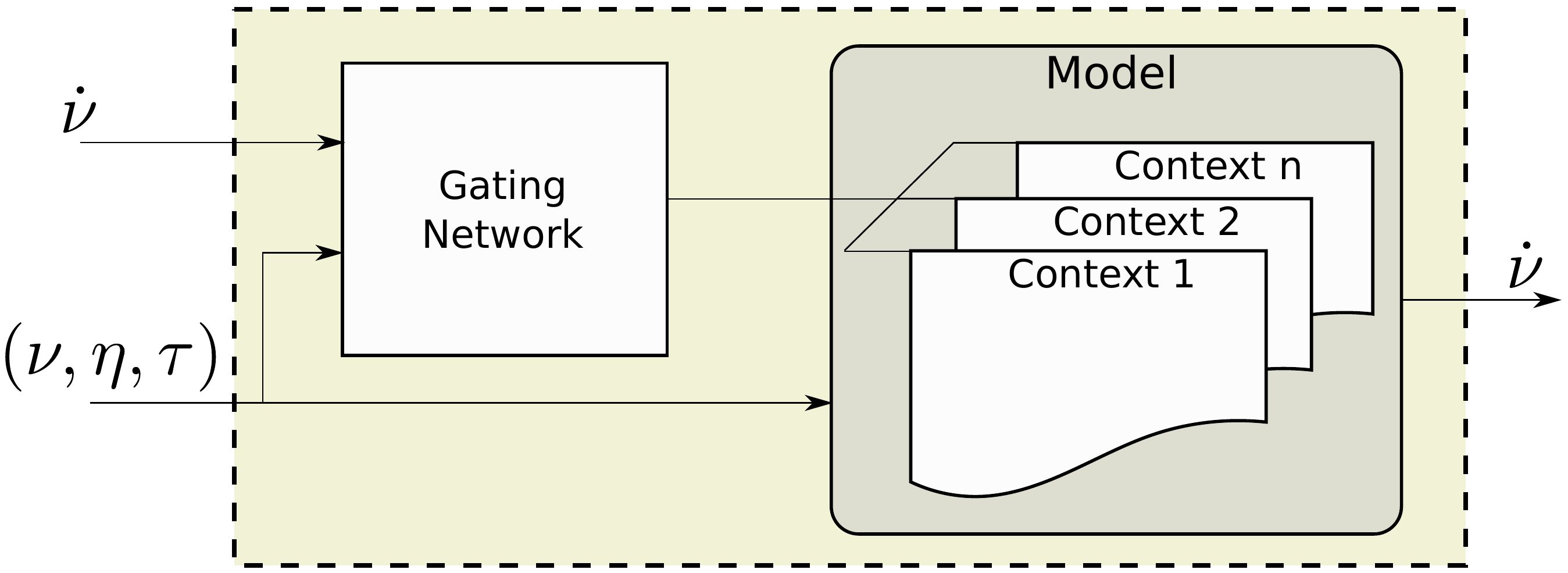}
\caption{The gating network classifier which acts as a decision layer on which
context to select.}
\label{fig:gating}
\vspace{-0.3cm}
\end{figure}

The goal of the gating network is to be able to infer from an observation
$\mathfrak{D} = \left\langle \bm{\dot{\nu}},\bm{\nu}, \bm{\eta},\bm{\tau} 
\right\rangle$, the true underlying model context currently in action. The gating network (Fig.~\ref{fig:gating})
can be seen as a decision layer that determines the active context. We implement
the gating network in this work as a multi-class classifier taking the state
and control input observations as a feature inputs, denoted as 
$Clf\left( \bm{\dot{\nu}},\bm{\nu}, \bm{\eta},\bm{\tau} \right)$. 

We differentiate between two types of gating networks, temporal and 
non-temporal. As their name infers, temporal learning algorithms can model
the temporal relations in the data if they exist, taking in data in the form of a time-series of shape 
($n$, $l$, $f$). Here, $n$ is the number of samples, $l$ is the series length (or time-steps),
and $f$ is the number of features. Non-temporal algorithms do not model any time
dependency in the data, and thus take a feature vector as input,
of shape ($N$, $f$), where in this case $N = n\times l$.

\subsection{Non-Temporal Learning}
We use three standard classifiers as our baseline namely SVM, MLP, and random forest
(RF) classifier. We won't consider any method
involving GPs in this study due to their cubic time complexity, which makes such
method impractical to train on a datasets larger than 10K samples.
In the following, we describe briefly the non-temporal baseline algorithms 
we use to classify the model contexts.
\subsubsection{Support Vector Machine}\label{sec:svm}
SVMs are binary classifiers that map the input vectors into a higher dimensional
feature space using a kernel transformation, and then fit a decision hyperplane
linearly onto this space \cite{Cortes1995}. For classifying multiple classes
with SVM, the problem is split into multiple binary classification problems. In
our implementation, we distinguish between every pair of classes, known as the
one-versus-one (ovo) approach, to avoid class imbalance problem that would come with the one-vs-all approach. As kernel, we use the radial-basis-function (RBF).
The hyperparameters to be optimized for the SVM are given as $\left( C, \gamma\right)$,
 where $C$ is a regularization parameter and $\gamma$ is the kernel's length scale.

\subsubsection{Multi-layer Perceptron}
The basic element of an MLP is a single neuron, which is a linear weighted sum of several inputs with a
non-linear activation function at its output \cite{alpaydin2014introduction}. Several neurons
can be stacked to form one layer, and thereafter several layers are connected
to form an MLP. MLPs are  trained using the back-propagation
method. To perform classification tasks, MLPs minimize a cross-entropy loss
function, with a softmax activation at the last layer to handle multi-class problems
\cite{alpaydin2014introduction}. For our application, we use a standard architecture
with three hidden fully-connected layers with a size of [256, 512, 128] respectively.
Every layer uses a ReLU \cite{glorot2011deep} activation function, with and a dropout
 \cite{srivastava2014dropout}
of 50\% to reduce overfitting. The output layer uses a softmax activation as mentioned earlier.
The total number of parameters used is 201,098.

\subsubsection{Random Forest}\label{sec:rf}
RF is an ensemble method which fits a set of decision trees classifiers by
randomly sampling from the training set with replacement. The decision of the RF
is computed by averaging the prediction of the individual classifiers 
\cite{breiman2001random}. Additionally, RFs can naturally handle multiple classes.
We optimize two hyperparameters for this method, namely the number of individual
trees used and the maximum depth of a each tree.

\subsection{Temporal Learning with LSTM Networks}
LSTM networks are recurrent neural networks used for modeling long-term
dependencies in time series. LSTMs avoid the vanishing gradient problem that
classical recurrent networks suffer from, by adding special types of forgetting
and remembering gates \cite{hochreiter1997long}.
LSTMs have a chain like structure with a repeating module. The repeating module
consists of four networks (also called gates) namely, input and output gates,
a forget-gate and a state-update-gate.
A detailed explanation can be found in \cite{hochreiter1997long}.

\begin{figure}[t!]
\centering
\includegraphics[trim=5 0 0 0,clip,width=0.47\textwidth]{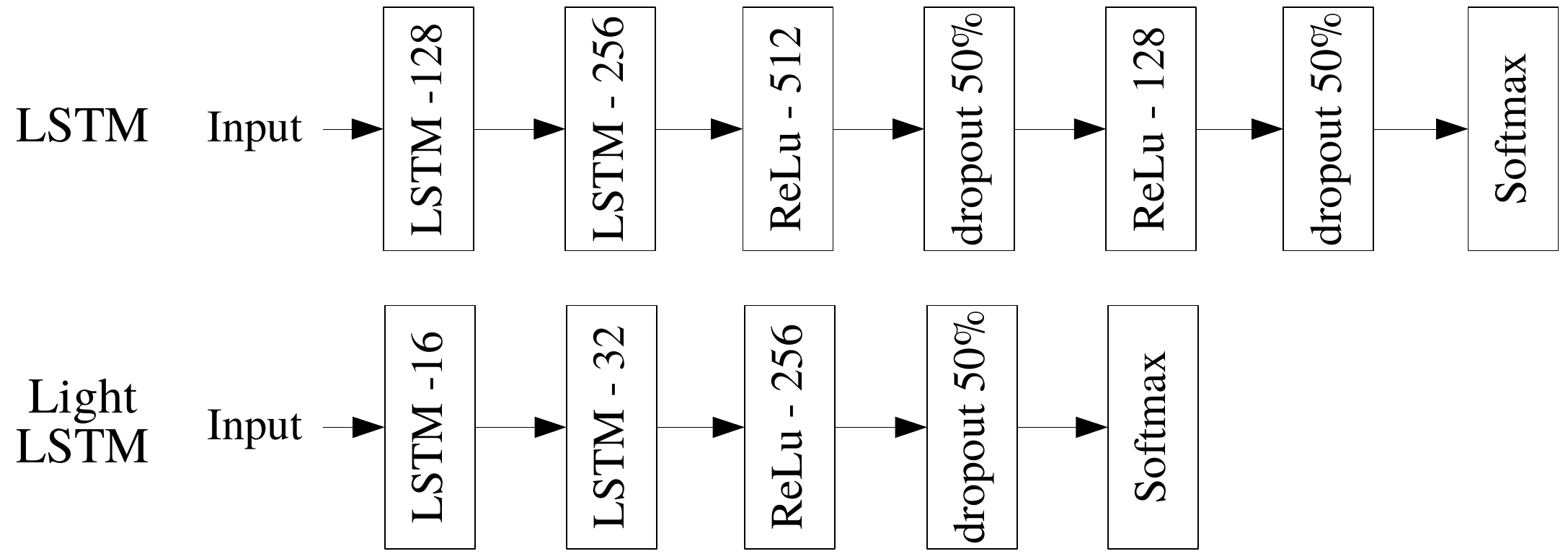}
\caption{Architectures of LSTM networks. {\bf Top:} standard LSTM network with two
LSTM layers, two Dense layers with dropouts and a softmax output layer. 
{\bf Bot.:} Light LSTM architecture with two small LSTM layers, one Dense layer 
and a softmax output.}
\label{fig:lstm}
\vspace{-0.35cm}
\end{figure}

We implement two architectures of LSTM networks Fig.~(\ref{fig:lstm}) as follows. The first network has a bigger architecture, with an LSTM input layer of 128 nodes, followed by another 
hidden LSTM layer with 256 nodes, then followed by two fully-connected
(dense) layers with 512 and 128 nodes, respectively. Both dense layers have a
dropout of 50\% each to avoid overfitting. The output is a softmax layer, and the
total number of parameters is 662,531.
We denote the second network as "Light LSTM" due to its smaller configuration,
including two LSTM layers with 16 and 32 nodes respectively, one dense layer with
256 nodes and 50\% dropout, and a softmax output layer. The overall number of parameters of light LSTM is 17,155.

\section{Evaluation and Results}
First, we briefly describe the robotic platform used for our experiment, and the
data collection procedure. Next, we describe the simulation model we fit using
the collected dataset, and the generation of the synthetic datasets. We report
afterwards, the training and model evaluation procedures, and compare
the testing results of each model.
\subsection{The Robotic Platform}
We use the AUV Dagon (Fig.~\ref{fig:dagon}) as a testing platform to collect the
data needed to fit the model described in Eq.~(\ref{dynamic}). Dagon
is a hovering type AUV that can be actuated in five DOFs (roll is passive).
A detailed description of the vehicle can be found in 
\cite{hildebrandt2010design}. To reduce the dimensionality of the problem, we
stabilize Dagon in the pitch and heave DOFs, and let it run freely in surge,
sway and yaw DOFs. We actuate the three lateral thrusters with a sinusoidal
signal of varying frequencies in order to cover as much range as possible of the
model's state space. A more detailed description of the data collection procedure
can be found in \cite{wehbe2017learning}. Using this dataset with a total of 2063
samples, we identify the parameters of Eq.~(\ref{model}) by minimizing the sum of
squared errors.

\subsection{Synthetic Data Generation}\label{sec:data}
We generate our datasets by running the resulting model through an
ordinary-differential-equation (ODE) solver. As an input signal, we give the
rotational velocity of each thruster in the form of a sinusoidal signal with
a period randomly selected between 20 and 70 seconds. For each model context,
we run the simulation for 40K seconds, randomly changing the value of the sine
periods every 1000 seconds. We sample the simulations with a frequency of 1 Hz
resulting in a dataset with 40K samples per context. The choice of these values was
made in order to have a large enough dataset that covers as much area as possible
of the model's state space. 
We assume the disturbances representing different contexts are in the robot
frame. By inducing disturbances in the simulation model, we generate data for
10 different model contexts described in Table~\ref{tab:contexts}. Disturbances in classes
1 to 4 were manually selected, whereas for the rest a we applied a random disturbance to the damping
and thrusters coefficients using a uniform distribution bounded to [0.5, 3]. We follow this approach
to ensure having physically realistic models that simulates random disturbances that might happen in a real
world scenario.

\begin{table}
\centering
 \caption{Description of different model contexts}
\label{tab:contexts}
\resizebox{0.3\textwidth}{!}{
\begin{tabular}{l c}
\hline
label     & context description \\ 
\hline
class 0  & nominal model   \\ 
class 1  & thruster 1 damage \\ 
class 2  & damping change - surge  \\ 
class 3  & thruster 1 broken \\ 
class 4  & damping change - yaw \\ 
class 5  & random damping change 1\\ 
class 6  & random damping change 2  \\ 
class 7  & random damping change 3 \\ 
class 8  & random thrusters configuration 1 \\ 
class 9  & random thrusters configuration 2 \\ 
\hline
\end{tabular}}
\end{table}

\subsection{Training the Gating Networks}
In this section, we describe the training procedure of the different methods. Moreover, we
study the effect of incrementally adding more classes on the performance
of each classifier. From the generated dataset described in Sec.~\ref{sec:data}
we construct three sets,
$\left\lbrace \mathfrak{D}^{\bm{3}}, \mathfrak{D}^{\bm{6}}, \mathfrak{D}^{\bm{10}}\right\rbrace$,
where $\mathfrak{D}^{\bm{3}}$ contains classes $0$ to $2$, $\mathfrak{D}^{\bm{6}}$
contains classes $0$ to $5$, and $\mathfrak{D}^{\bm{10}}$ contains all 10 classes.

\begin{figure*}[t!]
\centering
\subfloat[Grid search vs. no. of epochs length for LSTM]{\centering
\includegraphics[trim=10 10 0 5,clip,width=0.45\textwidth]{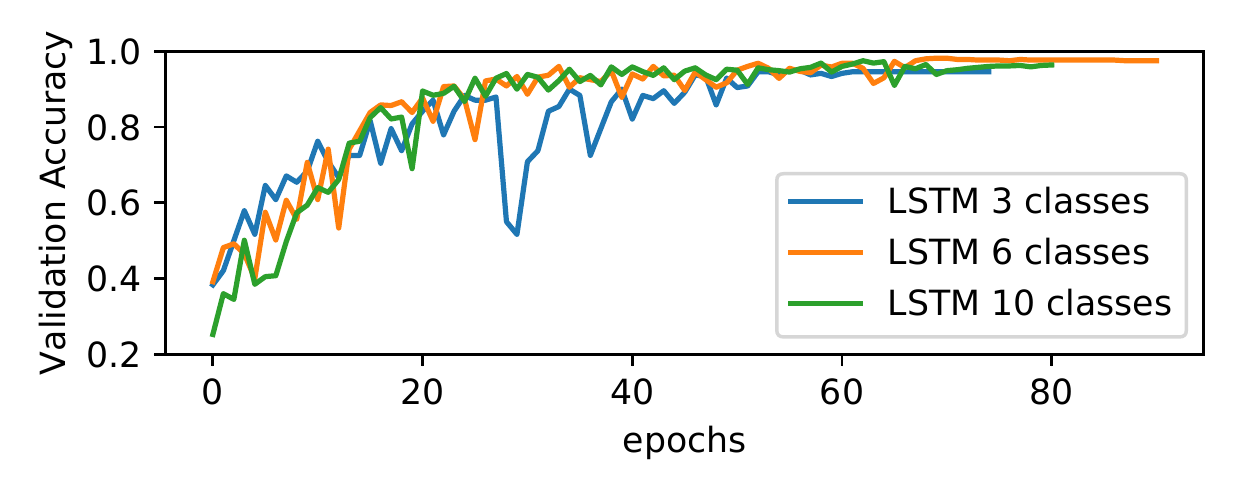}}
\subfloat[Grid search vs. no. of epochs length for Light LSTM]{\centering
\includegraphics[trim=0 10 10 5,clip,width=0.45\textwidth]{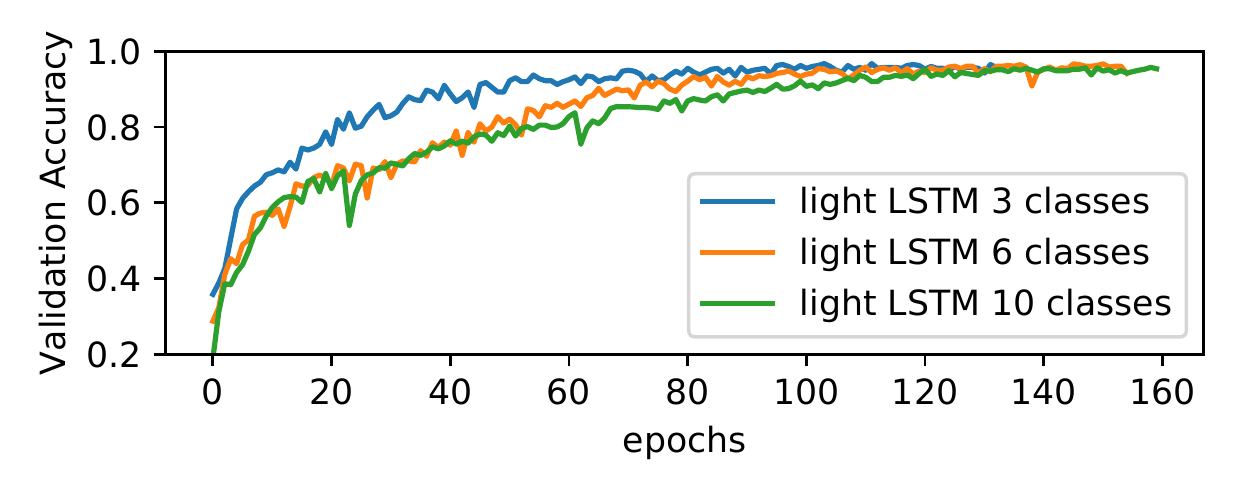}}
\caption{Validation accuracies with different epochs. A saturation in accuracy is
observed after 60 epochs for the LSTM network, whereas the Light LSTM required
around 120 epochs.}
\label{fig:epochs}
 \vspace{-0.2cm}
\end{figure*}

We split the generated datasets
into 3 consecutive (no shuffle) subsets: training, validation, and testing. The training and validation
sets are used to run a grid-search to find the best hyper parameters for the
classifiers. The classifier with the best validation accuracy is then evaluated on
the testing set. To keep the classes balanced, we use a stratified split with a
ratio of $60/20/20~\%$ for training, validation and testing respectively.
\begin{table}[t!]
\centering
 \caption{Cross-Validation results showing Best Validation Accuracies
  with the corresponding training accuracies}
\label{tab:validation}
\resizebox{0.45\textwidth}{!}{
\ \centering
 \begin{tabular}{c c c c c}
 \hline
 Classifier & dataset & 3 classes & 6 classes & 10 classes \\
 \hline
 \multirow{2}{*}{LSTM} 
       & training      & 0.966 & 0.998 &  0.998 \\
       & validation  &  0.946 & 0.977 & 0.973\\
\hline
 \multirow{2}{*}{Light LSTM} 
       & training    & 0.979 & 0.981 & 0.995 \\
       & validation & 0.947 & 0.956 & 0.964\\
\hline
 \multirow{2}{*}{SVM} 
       & training    & 0.854 & 0.844 &  0.847 \\
       & validation & 0.783 & 0.779 & 0.788\\
\hline
 \multirow{2}{*}{MLP} 
       & training     &0.784 &  0.751 & 0.688 \\
       & validation  &0.790 & 0.768 & 0.745\\
\hline
 \multirow{2}{*}{RF} 
       & training     & 0.970 & 0.936 & 0.867 \\
       & validation  & 0.578 & 0.553 & 0.504 \\
\hline
 \end{tabular}}
\end{table}
We train the LSTM networks using backpropagation-through-time method described in
\cite{hochreiter1997long}, and run several passes over the whole training set
(epochs). The learning rate is reduced by a factor of 0.9 after 10 epochs
if no improvement in the validation loss is observed. The training is stopped after
20 epochs with no improvement in the validation loss or if the improvement is less
than $10^{-4}$. We run this process as a grid-search for different time-series lengths.
Fig.~\ref{fig:epochs} reports the validation accuracy plotted against the number of
epochs for the two LSTM networks, where the standard LSTM requires less epochs to
converge than the ligh LSTM architecture. Although the time per epoch for the light
LSTM is less than that of the standard LSTM, the total time required by the light
version is slightly higher. 

\begin{figure*}[t!]
\centering
\subfloat [Grid search vs. time-series length for LSTM]{
\includegraphics[trim=10 10 0 5,clip,width=0.45\textwidth]{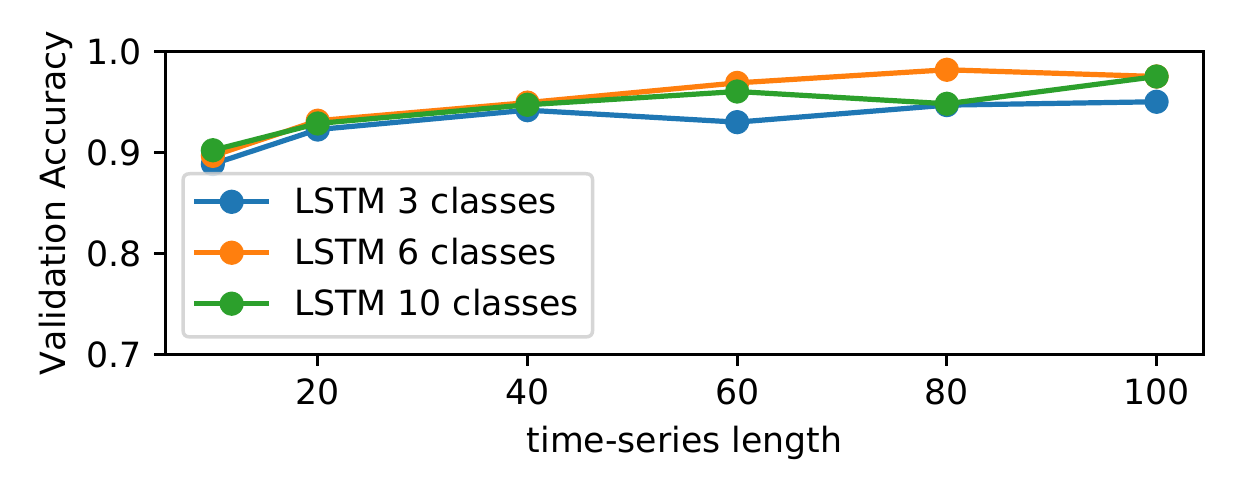}}
\subfloat[Grid search vs. time-series length for Light LSTM]{
\includegraphics[trim=0 10 10 5,clip,width=0.45\textwidth]{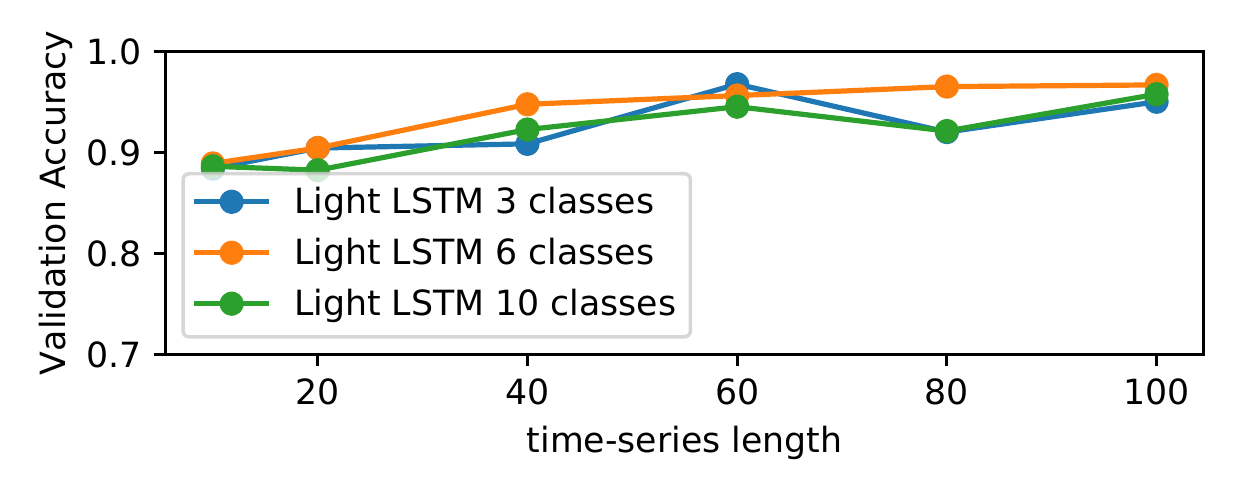}}
\caption{Validation accuracies LSTM with different series lengths. Results show
a saturation in accuracy of the LSTM for a series length of 100 samples. The Light LSTM shows
similar results except with the 3 classes dataset.}
\label{fig:validation}
 \vspace{-0.1cm}
\end{figure*}

In Fig.~\ref{fig:validation} we report the validation accuracies of both LSTMs
versus the length of the time-series. Both architectures achieve an accuracy
higher the $95\%$ with a time-series length of 80 and 100 samples.

We use the same mechanism to train the MLP network, except we only optimize for
the number of epochs since the MLP is a non-temporal method which takes in
data in the form of a feature vector. For the SVM and RF, we run a grid search
over the hyperparameters described in Sec.~\ref{sec:svm} \& \ref{sec:rf}.
Consequently, the hyperparameters resulting in the best validation accuracy
are selected for each classifier. Table~\ref{tab:validation} reports the best
validation accuracy for each method. We report also the corresponding training
accuracies to determine if a classifier overfits the data. An overfit can be
clearly noticed with the RF classifier, where the training accuracy is much
higher than the validation accuracy. Contrarily, all other methods show close
values of the training and validation accuracies. 

\begin{figure}[t!]
\centering
\includegraphics[trim=0 5 0 5,clip,width=0.42\textwidth]{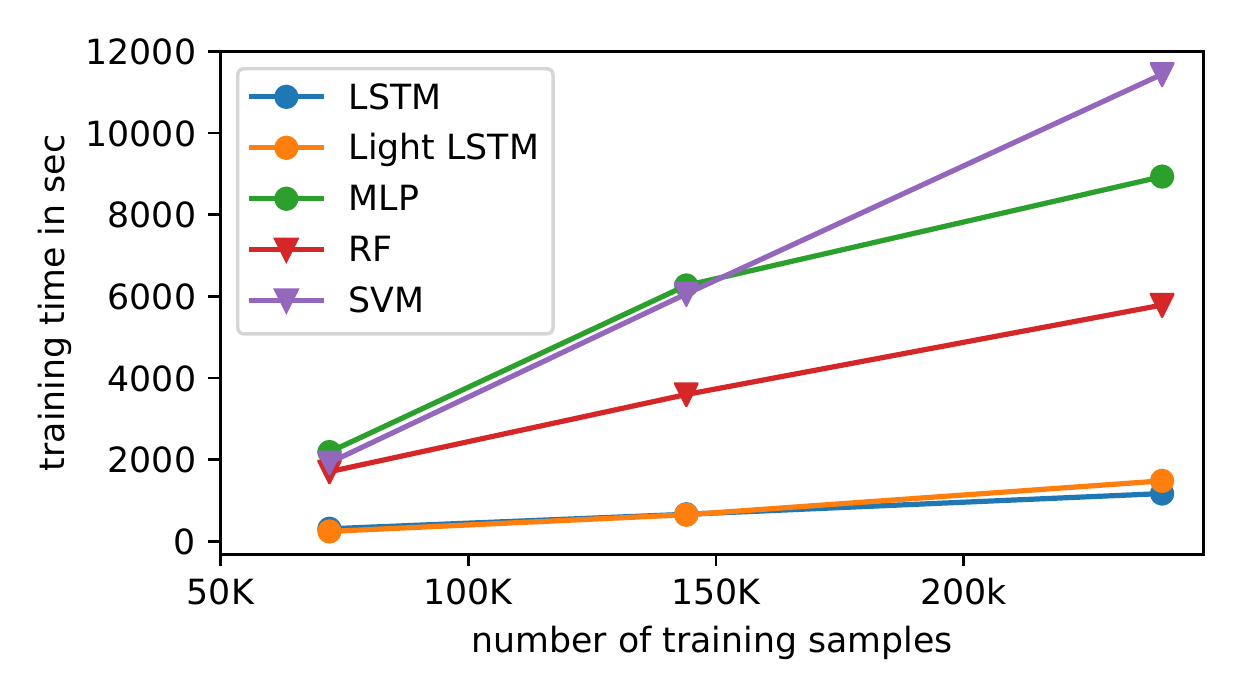}
\caption{Comparison of computational times for different classifiers used. Dot labels
represent methods trained on GPU and triangle labels represent methods trained on a CPU.}
\label{fig:train_time}
\vspace{-0.5cm}
\end{figure}

Furthermore, we compare the time complexity of the methods being evaluated.
According to \cite{hochreiter1997long}, an LSTM unit is local in space and time,
meaning that the time complexity does not depend on the network size and
the storage requirement is independent of the time-series length. These factors
render an LSTM to scale linearly with respect to the number of training samples.
On the other hand, the time complexity of an MLP with fully connected layers,
scales with the number number of neurons and hidden layers. Note that the LSTM
and MLP networks were trained on a GPU operating at 1 GHz, whereas the SVM and RF
where trained on a 10-core CPU operating at 3.3 GHz. To give a fair evaluation, we
only compare methods trained on the same kind of processor. The processing times are
depicted in in Fig.~\ref{fig:train_time}, where the LSTM networks show much more
computational efficiency as compared to the MLP. For example in the 10 classes
dataset, the MLP required an average of 39 seconds per epoch with 230 epochs to
reach the best validation accuracy, whereas the LSTM required only an average
of 16 seconds per epoch with 80 epochs for the validation accuracy to saturate.
On the other hand, the time complexity of RFs are proportional to the number of
trees in the forest, which in this case is still more efficient than the SVM which
scales quadratically with the number of samples.

\subsection{Test Results and Discussion}
\begin{figure}[t!]
\centering
\includegraphics[trim=0 5 0 0,clip,width=0.49\textwidth]{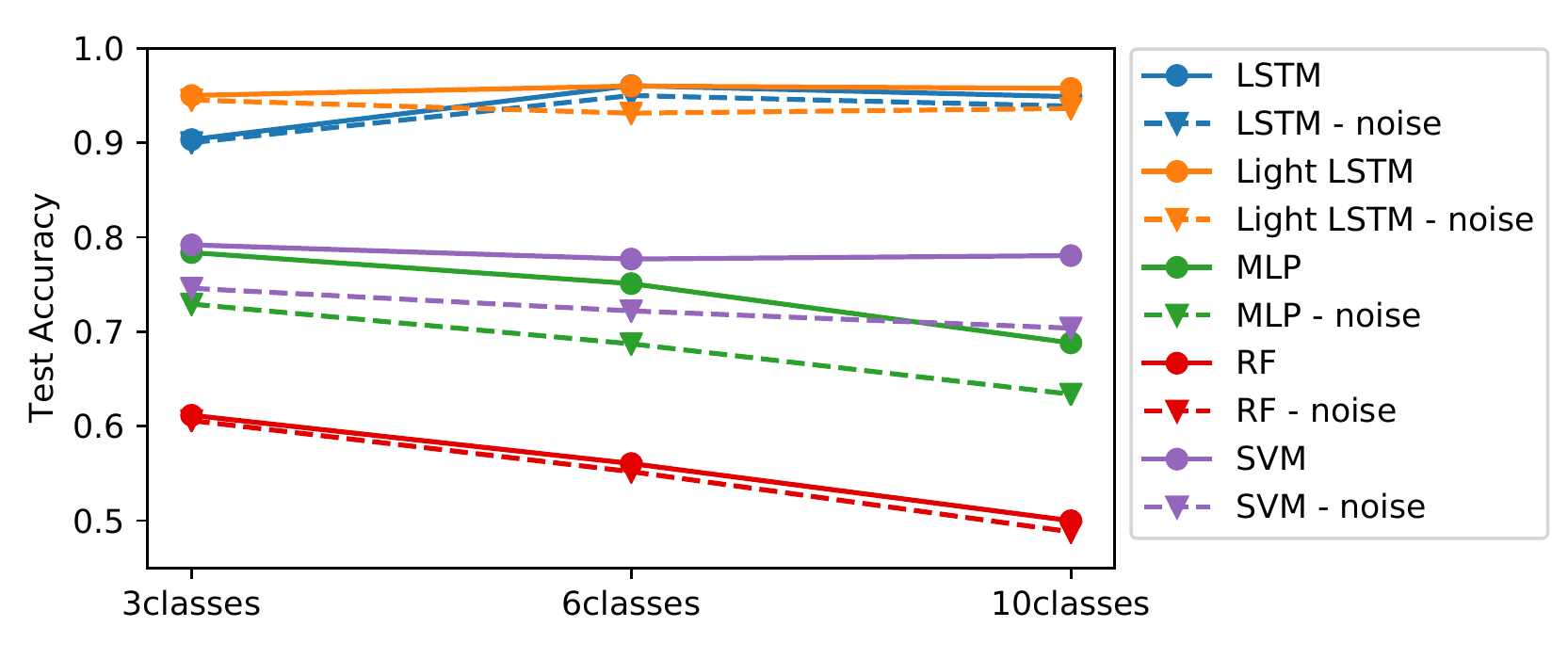}
\caption{Testing accuracies of the different classifiers. Dot labels represent noiseless data
and triangle labels represent noisy data.}
\label{fig:test}
\vspace{-0.2cm}
\end{figure}

After selecting the classifiers with the best performances on the validation set, we
evaluate how well can each classifier generalize on an unseen testing dataset that was
left completely independent from the hyperparameter optimization process.
Moreover, we test the effect of noise on the performance of the classifiers.
For this purpose we train the classifiers with the same datasets described, but
this time adding Gaussian noise similar to the natural noise observed on the
robot's sensors. 
The comparison of testing results is reported numerically in
Table~\ref{tab:test_acc} and graphically in Fig.~\ref{fig:test}. Both LSTM
networks show a prediction accuracy higher than 90$\%$ in all cases, whereas
none of the other baseline classifiers achieves more than 80$\%$ accuracy. With an increasing
number of classes and therefore increasing number of samples, the LSTM networks
maintain a consistent performance. It can also be noticed that the light LSTM network
performs even better than the larger LSTM for the 3 classes case. This result is not
surprising a large network tends more to overfit and less generalize on small training datasets.
Moreover, the SVM shows a consistent performance with increasing the number of
classes, whereas the performance of both the MLP and the RF degrades with additional
classes. A decrease of performance is usually expected because the classification problem
becomes more challenging with more classes. Eventually, with more classes
come with more data, and if a classifier can take advantage of that and obtain a better
overall understanding of the data, performance might increase.
\begin{table}
\centering
 \caption{Test Accuracies of different classifiers with the effect of adding noise}
\label{tab:test_acc}
\resizebox{0.45\textwidth}{!}{
\ \centering
 \begin{tabular}{c c c c c}
 \hline
 Classifier & noise & 3 classes & 6 classes & 10 classes \\
 \hline
 \multirow{2}{*}{LSTM} 
       & no noise & 0.903 & {\bf 0.960} & 0.949 \\
       & noise      &  0.900 & {\bf 0.950} & {\bf 0.939}\\
\hline
 \multirow{2}{*}{Light LSTM} 
       & no noise & {\bf 0.950} & {\bf 0.960} & {\bf 0.958} \\
       & noise     &  {\bf 0.945} & 0.931 & 0.936\\
\hline
 \multirow{2}{*}{SVM} 
       & no noise & 0.792 & 0.777 &  0.781 \\
       & noise & 0.746 & 0.722 & 0.704\\
\hline
 \multirow{2}{*}{MLP} 
       & no noise &0.784 &  0.751 & 0.688 \\
       & noise  &0.729 & 0.687 & 0.634\\
\hline
 \multirow{2}{*}{RF} 
       & no noise & 0.612 & 0.561 & 0.500 \\
       & noise  & 0.606 & 0.552 & 0.488 \\
\hline
 \end{tabular}}
 \vspace{-0.2cm}
\end{table}
In addition, we analyse the effect of noisy data on the performance of the methods
being tested. The LSTM networks show a high robustness against noise with only a
slight drop in the accuracy with the noisy datasets. The RF shows also an invariance
in the results when evaluated with noisy data, although its overall performance
is the worst amongst the other methods. On the contrary, the SVM and MLP
classifiers show a significant drop in performance with noisy data.

Finally, we present an extreme case where we generate 90 more classes on
top of the original $\mathfrak{D}^{\bm{10}}$ dataset, resulting in a dataset of 100 classes with
a total of 4 million samples. The additional classes were generated by inducing
random disturbances onto the damping and thrusters parameters, in a similar
fashion as classes 5 to 9. We evaluate only the LSTM networks on this dataset,
since training the other algorithms would have taken too much processing
resources. Table~\ref{tab:test_100} shows the test results, where both networks
maintain a very high classification accuracy (98-99$\%$). These results show clearly
the capability of LSTM networks to generalize with very high accuracy on datasets that are considered extremely
large in robotic applications.

\begin{table}[t!]
\centering
 \caption{Test Accuracies and training times of LSTMs with 100 classes}
\label{tab:test_100}
\resizebox{0.4\textwidth}{!}{
\ \centering
 \begin{tabular}{c c c c}
 \hline
 Classifier & noise & Test Accuracy & Training time \\
 \hline
 \multirow{2}{*}{LSTM} 
       & no noise & {\bf 0.994} & 12865 sec  \\
       & noise      & {\bf 0.992} & 12798 sec\\
\hline
 \multirow{2}{*}{Light LSTM} 
       & no noise & 0.988 &  14080 sec\\
       & noise     &  0.985  &  14400 sec\\
\hline
 \end{tabular}}
\vspace{-0.4cm}
\end{table}
\section{CONCLUSIONS}
In this work, we demonstrated the capability of LSTM networks to classify
accurately different contexts of an AUV model. The LSTM showed high robustness
when dealing with increasing number of classes as well as the ability to generalize
on noisy data, outperforming all other baseline classifiers tested in this paper.
Another advantage of LSTMs is their scalability on big datasets (up to 4M samples).
As future work, we aim to perform transfer learning with an LSTM network trained with
simulation onto real data.


\section*{ACKNOWLEDGMENT}
This work is part of the Europa-Explorer project (grant No. 50NA1704)
funded by the German Federal Ministry of Economics and Technology (BMWi).

\bibliographystyle{IEEEtran}
\bibliography{IEEEabrv,library}    

\end{document}